\documentclass{article}

\usepackage{arxiv}

\usepackage[utf8]{inputenc} %
\usepackage[T1]{fontenc}    %
\usepackage{hyperref}       %
\usepackage{url}            %
\usepackage{booktabs}       %
\usepackage{amsfonts}       %
\usepackage{nicefrac}       %
\usepackage{microtype}      %
\usepackage{lipsum}		%
\usepackage{graphicx}
\usepackage{natbib}
\usepackage{doi}

\usepackage{multirow}
\usepackage{comment}

\usepackage{titletoc}

\title{Revisiting Compositionality in Dual-Encoder Vision-Language Models: The Role of Inference}

\date{}	%

\author{Imanol Miranda\quad Ander Salaberria \quad Eneko Agirre \quad Gorka Azkune \\
HiTZ Center -- Ixa, University of the Basque Country (UPV/EHU) \\
\texttt{\{imanol.miranda, ander.salaberria, e.agirre, gorka.azcune\}@ehu.eus} \\
}

\renewcommand{\headeright}{}
\renewcommand{\undertitle}{}
\renewcommand{\shorttitle}{Revisiting Compositionality in Dual-Encoder Vision-Language Models: The Role of Inference}

\hypersetup{
pdftitle={Revisiting Compositionality in Dual-Encoder Vision-Language Models: The Role of Inference},
pdfsubject={},
pdfauthor={},
pdfkeywords={},
}

\begin{document}
\maketitle

\begin{abstract}
	Dual-encoder Vision-Language Models (VLMs) such as CLIP are often characterized as bag-of-words systems due to their poor performance on compositional benchmarks. We argue that this limitation may stem less from deficient representations than from the standard inference protocol based on global cosine similarity. First, through controlled diagnostic experiments, we show that explicitly enforcing fine-grained region–segment alignment at inference dramatically improves compositional performance without updating pretrained encoders. We then introduce a lightweight transformer that learns such alignments directly from frozen patch and token embeddings. Comparing against full fine-tuning and prior end-to-end compositional training methods, we find that although these approaches improve in-domain retrieval, their gains do not consistently transfer under distribution shift. In contrast, learning localized alignment over frozen representations matches full fine-tuning on in-domain retrieval while yielding substantial improvements on controlled out-of-domain compositional benchmarks. These results identify global embedding matching as a key bottleneck in dual-encoder VLMs and highlight the importance of alignment mechanisms for robust compositional generalization.  
\end{abstract}


\section{Introduction}
\label{sec:intro}
\begin{figure}
  \centering \includegraphics[width=0.6\textwidth]{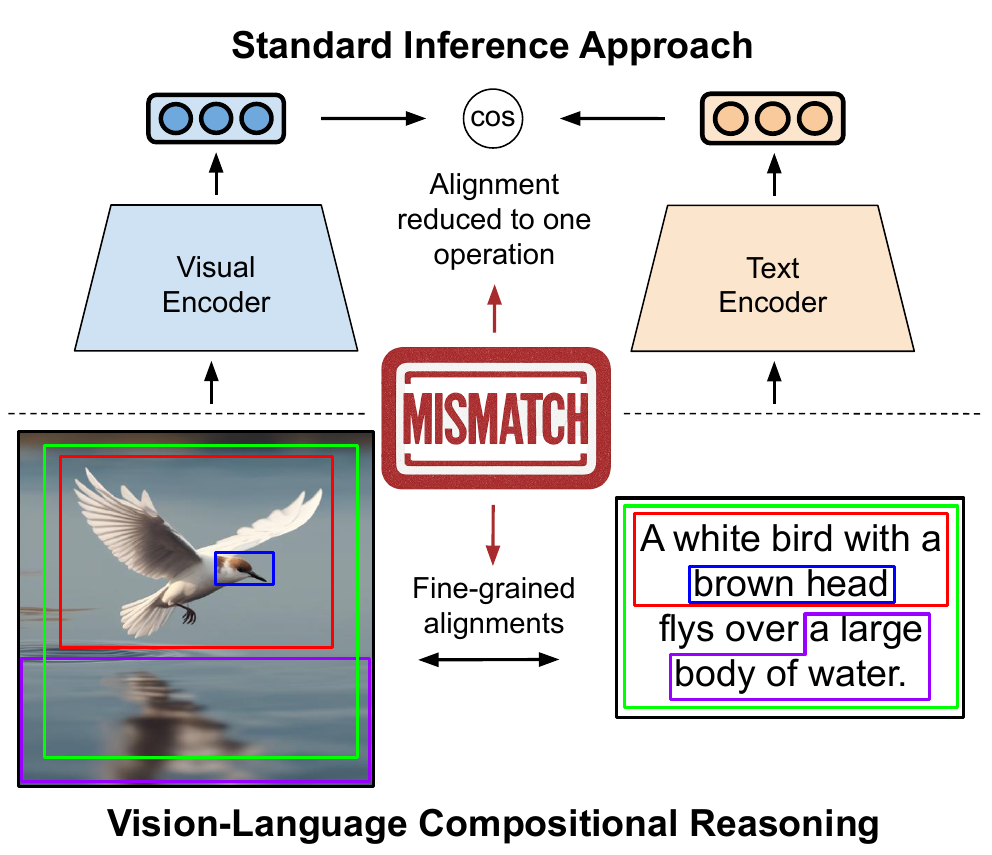}
  \caption{Vision-language compositional reasoning requires fine-grained alignment between textual segments describing objects, attributes, and relations and their corresponding image regions (bottom). In contrast, standard dual-encoder VLM inference relies on global embedding matching, reducing image–text similarity to a single cosine operation between pooled representations (top). This discrepancy highlights a structural mismatch between the localized alignment required for compositional binding and the global matching mechanism typically used at inference.}
  \label{fig:motivation}
\end{figure}

Dual-encoder Vision-Language Models (VLMs), pioneered by CLIP \citep{radford2021learning}, have become foundational models for vision-language tasks such as image–text retrieval \citep{cao2022image}, zero-shot image recognition \citep{radford2021learning}, and image–text scoring \citep{clipscore2021}. However, several studies report that these models struggle with compositional reasoning \citep{yuksekgonul2022and, thrush2022winoground, hsieh2024sugarcrepe, miranda2024BiVLC}, often behaving like bag-of-words systems. For instance, they fail to reliably distinguish between \emph{“a black dog and a white cat”} and \emph{“a black cat and a white dog”}, leading to low performance on Vision-Language Compositionality (VLC) benchmarks. At the same time, recent analyses caution against drawing definitive conclusions about the compositional limitations of VLMs \citep{diwan2022winoground, campbell2024understanding}.

We argue that these limitations may stem not from the learned representations themselves, but from the standard inference protocol used to evaluate them: global cosine similarity between pooled image and text embeddings. This global matching mechanism collapses multimodal information into a single vector comparison, potentially discarding region–token correspondences required for compositional binding. In contrast, compositional reasoning requires aligning textual elements describing objects, attributes, and relations with their corresponding image regions (Figure~\ref{fig:motivation}).

To test this hypothesis, we first conduct a controlled diagnostic study in which we enforce structured region–segment matching at inference time while keeping pretrained VLM encoders frozen. This experiment isolates the role of alignment from representation learning. We evaluate models on a new controlled out-of-distribution dataset, \textsc{BiSCoR-Ctrl}, specifically designed to measure compositional reasoning under reduced spurious correlations. Across multiple VLMs, enforcing structured alignment dramatically improves compositional performance, demonstrating that fine-grained region-segment alignments are beneficial for compositional reasoning.

Building on this diagnostic evidence, we then investigate whether such fine-grained alignment can be learned directly from pretrained representations. We introduce a lightweight transformer operating over patch and token embeddings extracted from frozen encoders. %
To disentangle alignment learning from representation updates, we compare against two controlled alternatives: (i) applying the same transformer to global embeddings only, and (ii) fully fine-tuning the original VLMs while retaining cosine-based inference. We further compare against prior end-to-end compositional training methods that modify the contrastive objective to encourage fine-grained reasoning. Models are evaluated on in-domain compositional benchmarks, including \textsc{SugarCrepe} \citep{hsieh2024sugarcrepe} and \textsc{BiVLC} \citep{miranda2024BiVLC}, as well as on \textsc{BiSCoR-Ctrl} for out-of-distribution generalization.

Our results show that:
\begin{enumerate}
\item Enforcing structured region–segment alignment at inference substantially improves compositional reasoning without updating pretrained encoders.
\item Learning fine-grained alignment over frozen patch and token representations matches full fine-tuning on in-domain retrieval and yields large gains in out-of-domain compositional generalization, outperforming cosine-based inference.
\item In contrast, prior end-to-end compositional training methods and full fine-tuning improve in-domain benchmarks but do not consistently transfer under distribution shift.
\item Increasing modeling capacity alone—e.g., applying a transformer to global embeddings—does not improve compositional performance, highlighting global embedding matching as a key bottleneck.
\end{enumerate}

Together, these findings suggest that compositional failures in dual-encoder VLMs stem less from missing representational capacity and more from the global embedding matching mechanism used at inference. Even models trained with compositional objectives show limited robustness under distribution shift when relying on global embedding similarity. By contrast, enabling localized alignment over pretrained representations leads to substantially stronger generalization on out-of-distribution benchmarks. These results highlight the importance of revisiting inference protocols in dual-encoder architectures and motivate evaluation settings that better reflect the structured, region–token information already encoded in modern VLMs. Code\footnote{https://github.com/IMirandaM/revisiting-vl-compositional-inference} and datasets\footnote{https://huggingface.co/datasets/imirandam/BiSCoR} will be publicly released.

\section{Related work}
\label{sec:sota}

\paragraph{Dual-encoder Vision-Language Models} consist of a vision encoder and a text encoder trained jointly to align visual and textual representations in a shared embedding space. The paradigm was popularized by CLIP~\citep{radford2021learning}, which enabled large-scale contrastive pretraining and zero-shot transfer. Subsequent models, including SigLIP and SigLIP 2~\citep{zhai2023sigmoid, tschannen2025siglip} and Perception Encoder~\citep{bolya2025perception}, refine the training objective, scale data and model size, and incorporate additional learning signals, consistently improving standard benchmarks such as image recognition and image–text retrieval.

Despite these differences, dual-encoder VLMs share a common inference mechanism: images and texts are independently encoded into global embeddings, and similarity is computed via cosine matching between pooled representations. This design underlies their success in retrieval, scoring~\citep{clipscore2021}, and as visual backbones for multimodal large language models such as LLaVA~\citep{liu2023visual}, PaliGemma~\citep{beyer2024paligemma}, and Qwen3-VL~\citep{bai2025qwen3}.

In this work, we examine whether this shared global matching paradigm is sufficient for assessing and exploiting compositional reasoning in dual-encoder VLMs.

\paragraph{Vision-Language Compositionality (VLC)} refers to a model’s ability to distinguish image-text pairs containing the same elements arranged in different configurations (e.g., “a red sphere and a blue cylinder” vs. “a blue sphere and a red cylinder”), requiring correct cross-modal binding of objects, attributes, and relations.

VLC is commonly evaluated through retrieval benchmarks that construct hard negatives by minimally perturbing captions or images~\citep{ma2023crepe, hsieh2024sugarcrepe, thrush2022winoground, ray2024cola}. Models must rank correct pairs above compositionally altered alternatives, typically using global cosine similarity between pooled embeddings. However, this setup allows models to exploit linguistic or dataset biases~\citep{lin2024evaluating, miranda2024BiVLC} and entangles compositional reasoning with the limitations of global matching.

To address these issues, recent work advocates bidirectional retrieval and controlled dataset design~\citep{miranda2024BiVLC}. We follow these practices and additionally introduce a controlled out-of-distribution benchmark to better isolate compositional reasoning from spurious correlations.

\paragraph{End-to-end training for VL compositionality.} The limited compositional performance of dual-encoder VLMs has motivated approaches that modify the training objective to encourage stronger cross-modal binding. One line of work introduces fine-grained alignment signals during pretraining, as in FILIP~\citep{lewei2021filip}, X-VLM~\citep{zeng2022multi}, PyramidCLIP~\citep{gao2022pyramidclip}, FSC-CLIP~\citep{oh2024preserving}, and FineCLIP~\citep{jing2024fineclip}, which promote token- or region-level interactions alongside global contrastive learning. Another direction emphasizes hard-negative mining to reduce shortcut learning, including NegCLIP~\citep{yuksekgonul2022and}, GNM~\citep{sahin2024enhancing}, and TripletCLIP~\citep{patel2024TripletCLIP}.

Despite differing training strategies, these methods retain the standard inference paradigm: image and text representations are pooled into global embeddings and compared via cosine similarity. In contrast, we investigate whether restructuring inference—while keeping pretrained encoders frozen—can better exploit fine-grained information already encoded in dual-encoder representations.

\section{Diagnosing the Inference Bottleneck in VLMs}
\label{sec:diagnosing}
Standard evaluations of dual-encoder VLMs rely on global cosine similarity between pooled image and text embeddings, without explicitly modeling interactions between image regions and textual components \citep{hsieh2024sugarcrepe, miranda2024BiVLC, yuksekgonul2022and}. Whether this global matching mechanism adequately captures the fine-grained correspondences required for compositional reasoning remains unclear. To investigate this question, we conduct a controlled diagnostic study that isolates the role of inference. We introduce \textsc{BiSCoR-Ctrl}, a benchmark designed to evaluate compositional robustness under controlled conditions, and apply a structured matching protocol to frozen VLM encoders at inference time. This setup allows us to assess whether enforcing region-segment alignment, without modifying the underlying representations, improves compositional performance.

\subsection{\textsc{BiSCoR-Ctrl}: A Controlled Compositional Benchmark}
\label{sec:biscor-ctrl}

Evaluating compositional reasoning in VLMs requires benchmarks that minimize confounding factors unrelated to visual grounding. Prior work shows that bidirectional retrieval settings provide a more balanced assessment of compositional abilities by mitigating language-based biases, where models may exploit caption plausibility or prior likelihood instead of image evidence \citep{lin2024revisiting, lin2024evaluating, miranda2024BiVLC, udandarao2025good}. Accordingly, \textsc{BiSCoR-Ctrl} is constructed as a symmetric bidirectional retrieval benchmark, requiring joint discrimination of images and captions under controlled conditions.

\textsc{BiSCoR-Ctrl} focuses exclusively on \textsc{Swap} cases, in which candidate captions contain the same objects, attributes, and relations but differ in their bindings (e.g., swapped attributes or object roles). Such cases are particularly challenging because success cannot rely on detecting individual elements in isolation \citep{miranda2024BiVLC, hsieh2024sugarcrepe, thrush2022winoground}. Instead, models must correctly bind attributes and relations to their corresponding objects. By restricting the dataset to this category, we focus on the core binding problem while eliminating easier compositional variations.

The benchmark is designed as an out-of-distribution (OOD) evaluation. It is built from CLEVR-based synthetic 3D scenes with controlled captions \citep{johnson2017clevr}, contrasting with the natural images and free-form text used in large-scale VLM pretraining \citep{radford2021learning, tschannen2025siglip, bolya2025perception}. The use of synthetic scenes enables precise control over object attributes and bindings, ensuring that positive and negative pairs differ only in the targeted compositional factor. Such control is essential for isolating whether models capture fine-grained compositional structure rather than dataset-specific correlations.

\paragraph{Dataset Construction:}

\begin{figure*}
  \centering \includegraphics[width=1\textwidth]{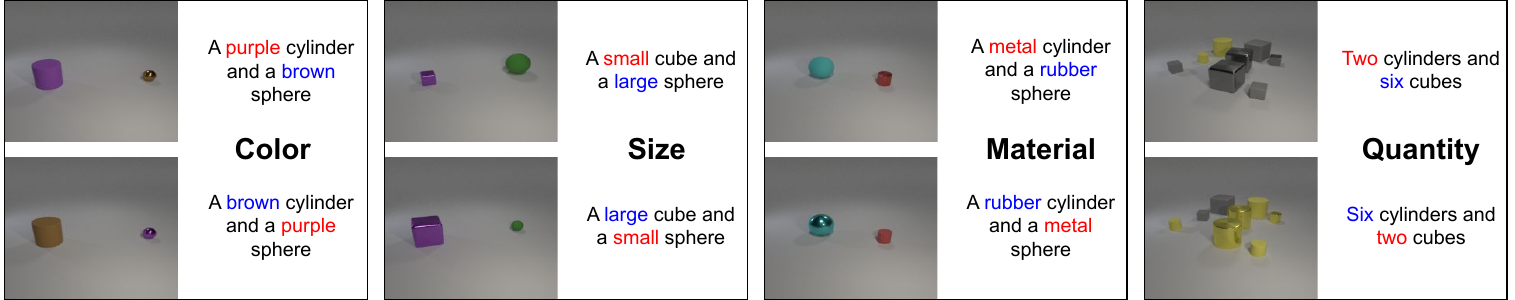}
  \caption{Examples from the \textsc{BiSCoR-Ctrl} dataset. From left to right: instances from the \textsc{Color}, \textsc{Size}, \textsc{Material}, and \textsc{Quantity} categories (the last containing 8 objects). Each instance consists of two image–caption pairs: a correct pair (top image and caption) and a hard negative pair (bottom image and caption).}
  \label{fig:BiSCoR-examples}
\end{figure*}

\textsc{BiSCoR-Ctrl} contains four compositional categories: \textsc{Color}, \textsc{Size}, \textsc{Material}, and \textsc{Quantity}, with 1,000 independent examples per category. Each example consists of two images and two captions forming a symmetric bidirectional retrieval pair: each caption correctly describes only one image, while the alternative caption corresponds to a minimally modified version of the same scene.

The dataset includes development and test splits of equal size, constructed from disjoint CLEVR scene sets (training split for development and validation split for test) to prevent scene-level overlap. Examples are generated automatically as follows:

\begin{enumerate}
    \item Scene Sampling. Sample a base CLEVR scene.
    \item Object Selection. For color, size, and material, select two objects; for quantity, select multiple objects sharing attributes to enable controlled counting variations.
    \item Caption Generation. Generate structured captions from predefined templates using scene annotations.
    \item Hard Negative Caption Creation (\textsc{Swap}). Create a compositional negative by swapping object identities or attribute bindings while preserving the same elements.
    \item Hard Negative Image Generation. Modify the original scene via controlled re-rendering, altering only the swapped attribute or binding. The resulting image pair preserves global layout and rendering conditions, differing only in the targeted compositional factor.
\end{enumerate}

This construction yields tightly controlled compositional contrasts in which captions share identical elements and images share the same global structure. Consequently, success on \textsc{BiSCoR-Ctrl} cannot rely on element detection or caption plausibility, but requires correctly binding object–attribute relationships across modalities. Example pairs are shown in Figure~\ref{fig:BiSCoR-examples} (see Appendix~\ref{appendix:dataset_info} for more).

\subsection{Enforcing Fine-Grained Alignment at Inference}
\label{subsec:sgi}

To assess whether explicit region–segment correspondences improve compositional reasoning, we design a controlled inference protocol that enforces such alignments while keeping pretrained VLM encoders frozen. The objective is diagnostic rather than methodological: by modifying only the inference procedure and leaving model parameters unchanged, we isolate the role of alignment independently of representation learning. Unlike standard evaluation based on global embedding similarity, our protocol decomposes captions into structured segments and matches them to image regions before aggregating similarities into a final score. We refer to this procedure as \textit{Structure-Guided Inference (SGI)}, described below.

\paragraph{[Step 1] Generate image crops:} Given an image $I$, we generate a set of crops $\{c_i\}_{i=0}^N$ using predefined scales, aspect ratios, and strides. These crops approximate candidate object-level regions without requiring supervision.

\paragraph{[Step 2] Generate text segments:} Given a caption $T$, we deterministically decompose it into object–attribute phrases plus the full caption using predefined parsing rules, obtaining segments $\{t_j\}_{j=0}^M$. For example, for $T=$ “a black cat and a white dog”, the segments are “black cat”, “white dog”, and the full caption. This decomposition is fixed across experiments.

\paragraph{[Step 3] Match crops and text segments:} Let $V_{enc}$ and $T_{enc}$ denote the frozen visual and textual encoders. For each crop $c_i$ and segment $t_j$, we compute embeddings $v_i = V_{enc}(c_i)$ and $l_j = T_{enc}(t_j)$ and their cosine similarity. This yields a similarity matrix of size $N \times M$. For each segment $t_j$, we select the crop with maximum similarity, producing one matched pair per segment $m_j$.

\paragraph{[Step 4] Aggregate alignment score:}
The final image–text score is computed as the average similarity over the selected matches:
\begin{equation}
Sim(I, T) = \frac{1}{M} \sum_j sim(m_j)    
\end{equation}

\subsection{Experimental Setup}

\paragraph{Models:} We evaluate representative dual-encoder VLMs: CLIP \citep{radford2021learning}, SigLIP 2 \citep{tschannen2025siglip}, and Perception Encoder (PE) \citep{bolya2025perception}, covering early contrastive models, modern multi-objective training, and recent scalable architectures. For CLIP we use ViT-B/32 ($224^2$ input); for SigLIP 2, ViT-B/32 and giant-opt-ViT/16 ($256^2$); for PE, ViT-B/16 ($224^2$) and ViT-G/14 ($448^2$). All models are evaluated using both standard global embedding similarity and the proposed Structure-Guided Inference (SGI). SGI hyperparameters (crop configuration and segment granularity) are selected on the \textsc{BiSCoR-Ctrl} development split and fixed thereafter (see Appendix~\ref{appendix:dev_results} for details).

Text segments are derived from \textsc{BiSCoR-Ctrl} scene annotations to ensure controlled decomposition. To verify robustness, we also evaluate automatically generated segments using SpaCy~\citep{spacy} (see Appendix\ref{appendix:sgi_spacy} for details), obtaining nearly identical results. Throughout the experiments, encoders remain frozen and no additional training is performed.

\paragraph{Benchmark and performance metrics:} For this diagnostic study, we use \textsc{BiSCoR-Ctrl} as the primary benchmark. Performance is measured with the group score \citep{thrush2022winoground}: a perfect group score requires both images to be correctly matched to their corresponding captions, and both captions to be correctly matched to their images (formal definitions are in Appendix~\ref{appendix:metrics}). This metric discourages solutions based solely on linguistic priors and provides a balanced assessment of compositional reasoning.

\subsection{Diagnostic Results and Analysis}
\label{subsec:sgi-results}
\begin{table}[t]
  \caption{Results for the diagnostic experiments on the \textsc{BiSCoR-Ctrl} dataset. We provide the average group score obtained by each model using global embedding similarity and SGI. We also provide the group score for all the categories of the dataset.}  
  \label{tab:esi-vs-sgi}
  \centering  
  \begin{tabular}{p{1.6cm}p{1.5cm}cccc}
    \toprule
    \multirow{2}{*}{\textbf{Model}} & \multirow{2}{1.5cm}{\centering \textbf{Average}} & \multicolumn{4}{c}{\textbf{Categorie}s} \\
     &  & \textsc{\textbf{Color}} & \textsc{\textbf{Size}} & \textsc{\textbf{Material}} & \textsc{\textbf{Quantity}} \\
    \midrule
    CLIP & \centering 1.4 & 1.5 & 0.2 & 0.7 & \textbf{3.3} \\
    \quad $\mathbf{+}$SGI & \centering\textbf{24.9} & \textbf{76.5} & \textbf{4.7} & \centering\textbf{15.7} & 2.5 \\
    \midrule
    PE & \centering8.5 & 5.9 & 10.9 & 7.7 & 9.4 \\
    \quad $\mathbf{+}$SGI & \centering\textbf{44.7} & \textbf{91.3} & \textbf{16.9} & \textbf{55.5} & \textbf{15.3} \\
    \midrule
    PE-G & \centering 3.8 & 5.8 & 1.0 & 2.6 &  5.8  \\
    \quad $\mathbf{+}$SGI & \centering\textbf{52.9} & \textbf{97.4} & \textbf{6.7} & \textbf{92.1} & \textbf{15.5} \\
    \midrule
    SigLIP 2 & \centering4.5 & 3.7 & \textbf{9.6} & 1.5 & 3.1 \\
    \quad $\mathbf{+}$SGI & \centering\textbf{47.6} & \textbf{94.9} & 7.1 & \textbf{68.3} & \textbf{20.2} \\
    \midrule
    SigLIP 2-G & \centering4.8 & 6.0 & 6.6 & 2.4 & 4.4 \\
    \quad $\mathbf{+}$SGI & \centering\textbf{56.4} & \textbf{96.2} & \textbf{18.1} & \textbf{91.7} & \textbf{19.7} \\
    \bottomrule
  \end{tabular}
\end{table}

Table~\ref{tab:esi-vs-sgi} reports the results on \textsc{BiSCoR-Ctrl}. Using standard global cosine similarity, all models obtain extremely low group scores (1–9 points on average), confirming that global embedding matching fails to capture compositional structure. Enforcing fine-grained region–segment alignment at inference through SGI substantially improves performance, raising average group scores to 25–56 points. Gains are consistent across compositional categories, with color and material swaps benefiting most, while size and quantity remain more challenging.

Importantly, SGI operates on frozen encoders and introduces no additional training, modifying only the inference procedure. Similar results obtained with automatically generated text segments (see Appendix~\ref{appendix:sgi_spacy}) indicate that improvements stem from better exploitation of pretrained representations rather than privileged annotation.

Overall, these experiments indicate that explicitly enforcing fine-grained region–segment alignment at inference substantially improves compositional performance compared to standard global cosine similarity. However, because SGI relies on externally guided structural decomposition, it does not establish whether such alignment can emerge directly from pretrained representations. The results instead suggest that frozen VLM embeddings contain information that can support compositional alignment when appropriately structured. We therefore ask whether this alignment can be learned directly from patch- and token-level embeddings, without imposing external structure at inference time.

\section{Learning Fine-Grained Alignment from Frozen VLMs}
\label{sec:learning}

\subsection{Motivation and Overview}

Building on the diagnostic findings of Section~\ref{sec:diagnosing}, we investigate whether fine-grained alignment can be learned directly from pretrained VLM representations. Our hypothesis is that patch-level visual embeddings and token-level textual embeddings already encode localized information relevant for compositional reasoning, but lack an explicit mechanism to align these elements in a structured manner. To this end, we introduce a lightweight transformer that operates on frozen patch and token embeddings, learning cross-modal correspondences while keeping the pretrained VLM intact.

This design isolates the role of alignment learning from representation learning. Unlike full fine-tuning, which entangles both effects, our approach explicitly tests whether compositional improvements can be obtained by learning to exploit existing representations. If successful, this would provide further evidence that compositional reasoning in VLMs is primarily limited by alignment rather than encoding.

\subsection{Alignment Transformer Architecture}
To test whether compositional improvements can be obtained by learning fine-grained correspondences, we introduce a lightweight alignment module operating on top of frozen VLM encoders. The visual and textual backbones remain unchanged throughout training.

\paragraph{Frozen Patch and Token Embeddings:} Given an image $I$ and a caption $T$, we extract patch embeddings $\{ v_i\}_{i=1}^{N}$ from the visual encoder and token embeddings $\{ t_i\}_{i=1}^{M}$ from the text encoder. We discard the projection heads used for standard embedding similarity inference. All encoder parameters are frozen.

\paragraph{Cross-Modal Alignment Transformer:} We concatenate the visual and textual sequences and feed them into a lightweight transformer encoder. We learn new image and text projections on top of those embeddings. We also add learnable positional embeddings to all inputs, an additional embedding to separate visual and textual representations, and a \texttt{[CLS]} token to the transformer encoder to learn the final matching score between the image and the text.

The transformer allows: (i) cross-modal attention between patches and tokens, (ii) contextual refinement of region-level and token-level features, (iii) implicit learning of fine-grained correspondences. Importantly, the number of parameters in this module is small compared to the frozen backbone, ensuring that improvements can be attributed to alignment learning rather than large-scale representation updates.

\paragraph{Training Objective:} The alignment transformer is trained using a contrastive retrieval objective. Given a batch of image–text pairs, the model is optimized to assign higher scores to matching pairs than to mismatched ones. Crucially, only the alignment transformer parameters are updated; the visual and textual encoders remain frozen.

\paragraph{Design Rationale:} This architecture isolates the effect of alignment learning: (i) if improvements arise from full fine-tuning, compositional gains should only appear when updating the encoders, and (ii) if pretrained representations already encode compositional cues, learning to align patch- and token-level features should suffice. By keeping the backbone fixed and introducing only a lightweight cross-modal module, we directly test whether fine-grained alignment is the missing ingredient for compositional reasoning in VLMs.

\subsection{Experimental Setup}

\paragraph{Models:} We conduct experiments using CLIP \citep{radford2021learning} and Perception Encoder (PE) \citep{bolya2025perception}. These models were selected to cover both a widely studied contrastive VLM and a more recent large-scale alternative. %
For both CLIP and PE, we use the same pretrained checkpoints as in Section~\ref{sec:diagnosing}, discarding PE-G. In all alignment-learning experiments, the visual and textual encoders remain frozen.%

\paragraph{Training Data:} To study how data properties affect alignment learning, we consider two training datasets: (i) COCO \citep{lin2014microsoft}: a standard image–caption dataset containing clean positive pairs but no explicit hard negatives, and (ii) TROHN-Img \citep{miranda2024BiVLC}: a dataset derived from COCO that introduces automatically generated hard negative captions and images, resulting in a substantially noisier training distribution. This setup allows us to investigate whether exposure to hard negatives facilitates the learning of fine-grained correspondences, or whether alignment learning primarily depends on the structure of pretrained representations rather than negative sampling difficulty.

\paragraph{Evaluation Data:} We evaluate both in-domain and out-of-domain compositional reasoning. In-domain benchmarks: (i) \textsc{BiVLC} \citep{miranda2024BiVLC}: a bidirectional retrieval benchmark containing \textsc{Replace}, \textsc{Swap}, and \textsc{Add} compositional categories, and (ii) \textsc{SugarCrepe} \citep{hsieh2024sugarcrepe}: an image-to-text retrieval benchmark with the same three compositional transformations. These datasets test whether alignment learning improves compositional reasoning under training-distribution conditions. Note that both datasets are derived from COCO and thus are in-domain relative to our training datasets. Out-of-domain benchmark: \textsc{BiSCoR-Ctrl}, a controlled diagnostic benchmark designed to isolate compositional reasoning under distribution shifts (Section~\ref{sec:biscor-ctrl}). Evaluation on \textsc{BiSCoR-Ctrl} assesses whether learned alignments generalize beyond the training data.

\paragraph{Ablation Design:} To isolate the role of alignment learning, we compare the proposed lightweight alignment transformer with patch-token embeddings with two other training settings: (i) Full fine-tuning: the original dual-encoder VLM is fine-tuned end-to-end (visual and textual encoders) on COCO or TROHN-Img using a standard contrastive retrieval objective, without any additional alignment transformer. This setting entangles representation learning and alignment implicitly through encoder updates. (ii) Global-alignment transformer: the same lightweight transformer is trained on top of frozen global image and text embeddings only. This ablation controls for the additional modeling capacity of the transformer while removing access to patch- and token-level representations. 

This controlled comparison enables us to answer two key questions: (i) Are compositional gains primarily driven by updating representations? (ii) Or can they be achieved by learning to align existing fine-grained features?

\paragraph{Implementation Details and Hyperparameter Selection:} For the lightweight alignment transformer, we select architectural hyperparameters using development splits of the training datasets. In particular, we explore the number of transformer layers under each training configuration (COCO and TROHN-Img, global vs. fine-grained representations). For CLIP, we perform an exhaustive search over depths ranging from 1 to 4 layers (less than 10\% of the total parameters). The optimal number of layers is 4 for all the variants (see full results in Appendix~\ref{appendix:layer_setup}). This configuration contains 13.3M parameters, corresponding to only 8.8\% of the parameters of the original CLIP backbone, highlighting the lightweight nature of the alignment module. For PE, to reduce computational cost, we evaluate only the configurations that performed best for CLIP (4 layers). This controlled strategy allows us to assess whether alignment learning generalizes across backbones without repeating the full hyperparameter search. All hyperparameters are fixed before final evaluation on the test sets (see Appendix~\ref{appendix:learning_setup} and \ref{appendix:implementation}).

\subsection{Results and Discussion}
\label{subsec:main-results}
\begin{table}[t]
  \caption{Main results comparing full fine-tuning (FT), a transformer over frozen global embeddings (TF\textsubscript{Global}), and a transformer over frozen patch/token embeddings (TF\textsubscript{Local}). Results are reported as accuracy on \textsc{SugarCrepe} and group score on bidirectional retrieval benchmarks (\textsc{BiVLC}, \textsc{BiSCoR-Ctrl}); \textsc{Swap} performance for \textsc{SugarCrepe} and \textsc{BiVLC} is shown in brackets. While FT and TF\textsubscript{Local} improve in-domain performance, TF\textsubscript{Local} yields markedly stronger gains under distribution shift.}  
  \label{tab:alignment-learning}
  \centering  
  \begin{tabular}{lllccc}%
    \toprule
    \multirow{2}{*}{\textbf{Backbone}} & \multirow{2}{*}{\textbf{Method}} & \textbf{Training}  & \multicolumn{2}{c}{\textbf{In-domain}} & \textbf{Out-of-domain} \\
     & & \textbf{data} & \textbf{\textsc{SugarCrepe}} & \textbf{\textsc{BiVLC}} & \textbf{\textsc{BiSCoR-Ctrl}} \\
    \midrule
    \multirow{8}{*}{CLIP} & Base & Pretrained & 73.0\textsubscript{(63.3)} & 36.8\textsubscript{(13.7)} & 1.4 \\
    \cmidrule{2-6}
    &\multirow{2}{*}{FT} & COCO & 81.0\textsubscript{(68.9)} & 47.5\textsubscript{(20.9)}  & 1.4 \\
     && TROHN-Img & \underline{85.5}\textsubscript{(72.0)} & \underline{57.5}\textsubscript{(27.9)} & 1.9 \\
     \cmidrule{2-6}
    &\multirow{2}{*}{TF\textsubscript{Global}} & COCO & 73.2\textsubscript{(63.5)} & 38.4\textsubscript{(11.4)} & 1.2 \\
     && TROHN-Img & 82.2\textsubscript{(68.8)} & 49.0\textsubscript{(15.9)} & 1.2 \\
     \cmidrule{2-6}
    &\multirow{2}{*}{TF\textsubscript{Local}} & COCO & 80.9\textsubscript{(76.3)} & 45.7\textsubscript{(24.0)} & \textbf{15.1} \\
     && TROHN-Img & \textbf{86.3}\textsubscript{(77.1)} & \textbf{61.3}\textsubscript{(39.0)} & \underline{13.2} \\
    \midrule
    \multirow{8}{*}{PE} & \centering Base & Pretrained & 84.4\textsubscript{(76.8)} & 41.5\textsubscript{(13.4)}  & 8.5 \\
    \cmidrule{2-6}
    &\multirow{2}{*}{FT} & COCO & 85.7\textsubscript{(76.1)} & 56.3\textsubscript{(31.8)} & 8.3 \\
     && TROHN-Img & \textbf{90.1}\textsubscript{(80.3)} & \textbf{68.2}\textsubscript{(43.7)} & 9.8 \\
     \cmidrule{2-6}
    &\multirow{2}{*}{TF\textsubscript{Global}} & COCO & 78.0\textsubscript{(68.0)} & 44.6\textsubscript{(15.6)} & 1.9 \\
     && TROHN-Img & 86.9\textsubscript{(74.9)} & 58.3\textsubscript{(24.5)} & 1.2 \\
     \cmidrule{2-6}
    &\multirow{2}{*}{TF\textsubscript{Local}} & COCO & 84.4\textsubscript{(80.6)} & 53.3\textsubscript{(33.7)} & \textbf{30.0} \\
     && TROHN-Img & \underline{89.2}\textsubscript{(80.8)} & \underline{67.1}\textsubscript{(42.9)} & \underline{24.0} \\
    \bottomrule
  \end{tabular}
  \vspace{-0.5em}
\end{table}

Table~\ref{tab:alignment-learning} presents the main results (results per compositional category in Appendix~\ref{appendix:detailedlearning}). We analyze in-domain retrieval performance first, followed by out-of-domain compositional generalization.

\paragraph{In-Domain Retrieval:} Fine-tuning substantially improves performance on both \textsc{SugarCrepe} and \textsc{BiVLC}. For instance, CLIP fine-tuned on TROHN-Img increases from 73.0 to 85.5 on \textsc{SugarCrepe} and from 36.8 to 57.5 on \textsc{BiVLC}. Similar trends are observed for PE. These results confirm that standard fine-tuning is effective for improving retrieval under the training distribution. The lightweight alignment transformer operating on global representations (TF\textsubscript{Global}) yields modest gains in some in-domain settings, indicating that additional modeling capacity alone is insufficient.

When operating on patch- and token-level representations (TF\textsubscript{Local}), the alignment transformer matches or exceeds full fine-tuning performance. Notably, CLIP$\mathbf{+}$TF\textsubscript{Local} trained on TROHN-Img achieves the strongest in-domain results overall (86.3 on \textsc{SugarCrepe} and 61.3 on \textsc{BiVLC}), despite keeping the backbone frozen.

Since our hypothesis concerns compositional binding, we additionally analyze the \textsc{Swap} subsets of \textsc{SugarCrepe} and \textsc{BiVLC} (Table~\ref{tab:alignment-learning}, brackets), which isolate minimal attribute–object reversals and reduce reliance on linguistic shortcuts. In these subsets, the relative performance gaps between TF\textsubscript{Local}, fine-tuning, and TF\textsubscript{Global} become even more pronounced across both backbones, with TF\textsubscript{Local} consistently achieving the strongest results. This amplification on the most binding-sensitive cases reinforces that localized alignment particularly benefits examples requiring precise cross-modal correspondence. %

\paragraph{Out-of-Domain Compositional Generalization:} A strikingly different picture emerges on \textsc{BiSCoR-Ctrl}. Full fine-tuning provides negligible improvements over the frozen baseline (e.g., CLIP: 1.4 → 1.9; PE: 8.5 → 9.8). Similarly, the global transformer fails to improve compositional generalization and in some cases reduces it. In contrast, the fine-grained alignment transformer yields substantial gains. CLIP$\mathbf{+}$TF\textsubscript{Local} improves from 1.4 to 15.1 (COCO training), while PE$\mathbf{+}$TF\textsubscript{Local} improves from 8.5 to 30.0 — nearly a fourfold increase over the pretrained baseline. These improvements are obtained without updating the visual or textual encoders. This result provides strong evidence that compositional reasoning benefits primarily from learning fine-grained alignments rather than from updating global representations.

\paragraph{Alignment vs. Representation Learning:} The controlled comparison between full fine-tuning and frozen-backbone alignment learning allows us to disentangle representation learning from alignment learning. While fine-tuning improves in-domain retrieval, it fails to enhance compositional generalization. Conversely, learning to align patch- and token-level representations significantly improves out-of-domain compositional performance. These findings suggest that pretrained VLMs already encode rich localized information, but standard global objectives fail to exploit it. Alignment learning is sufficient to unlock this latent compositional capability.

\paragraph{Effect of Hard Negatives:} Training with TROHN-Img consistently improves in-domain retrieval across all methods. However, for fine-grained alignment learning, COCO-trained models often achieve stronger out-of-domain compositional performance (e.g., PE$\mathbf{+}$TF\textsubscript{Local}: 30.0 vs 24.0). This suggests that excessive noise or automatically generated hard negatives may bias the model toward shortcut strategies, reducing robustness under distribution shift.

\paragraph{Summary:} Overall, the results demonstrate that: (i) Fine-tuning improves in-domain retrieval but not compositional generalization; (ii) Additional modeling capacity alone is insufficient; (iii) Learning fine-grained alignment over frozen representations dramatically enhances compositional reasoning. Together with the diagnostic findings in Section~\ref{sec:diagnosing}, these results support our central claim: compositional failures in VLMs stem primarily from insufficient alignment mechanisms rather than missing representational capacity.

\subsection{Comparison with End-to-End Compositional Training}
\label{sec:end-to-end}

\begin{table}[t]  
  \caption{Comparison between end-to-end compositional training methods and alignment learning. We report accuracy on \textsc{SugarCrepe} and group score on \textsc{BiVLC} and \textsc{BiSCoR-Ctrl}; \textsc{Swap} results for \textsc{SugarCrepe} and \textsc{BiVLC} are shown in brackets. Training data is indicated for each method ($\dagger$ denotes dataset extensions with hard negatives or similarity-based sampling). While several end-to-end approaches improve in-domain benchmarks, gains do not consistently transfer to \textsc{BiSCoR-Ctrl}, whereas alignment learning over frozen representations yields stronger improvements under distribution shift.}

  \label{tab:end-to-end}
  \centering  
  \begin{tabular}{llccc}%
    \toprule    
     \textbf{Model} & \textbf{Training data} & \textbf{\textsc{SugarCrepe}} & \textbf{\textsc{BiVLC}} & \textbf{\textsc{BiSCoR-Ctrl}} \\
    \midrule
    CLIP & Pretrained & 73.0\textsubscript{(63.3)} & 36.8\textsubscript{(13.7)} & 1.4 \\
    \midrule
    NegCLIP & COCO$\dagger$ & 83.6\textsubscript{(76.6)} & 44.9\textsubscript{(18.7)} & 1.8 \\ %
    TripletCLIP & CC3M/12M & 82.6\textsubscript{(71.9)} & 35.2\textsubscript{(9.8)} &  1.2 \\ %
    FSC-CLIP & COCO$\dagger$ & \underline{85.1}\textsubscript{(77.6)} & \underline{46.5}\textsubscript{(19.2)} &  1.2 \\ %
    X-VLM & COCO$\mathbf{+}$Others & 81.9\textsubscript{(67.4)} & 40.9\textsubscript{(12.3)} & 1.7 \\ %
    FineCLIP & COCO$\dagger$ & 80.6\textsubscript{(68.2)} & 39.4\textsubscript{(8.6)} & 1.4 \\ %
    \midrule
    \multirow{2}{*}{CLIP$\mathbf{+}$TF\textsubscript{Local}} & COCO & 80.9\textsubscript{(76.3)} & 45.7\textsubscript{(24.0)} & \textbf{15.1} \\    
    & TROHN-Img & \textbf{86.3}\textsubscript{(77.1)} & \textbf{61.3}\textsubscript{(39.0)} & \underline{13.2} \\
    \bottomrule
  \end{tabular}
\end{table}

Recent work has sought to improve compositional reasoning by modifying CLIP-style training objectives, either through fine-grained alignment supervision (e.g., FSC-CLIP \citep{oh2024preserving}, X-VLM \citep{zeng2022multi}) or by introducing hard negatives (e.g., NegCLIP \citep{yuksekgonul2022and}, TripletCLIP \citep{patel2024TripletCLIP}). These methods alter the training signal but retain standard global cosine similarity at inference.

Table~\ref{tab:end-to-end} compares these approaches with our alignment-learning model built on frozen CLIP representations (detailed results per category in Appendix~\ref{appendix:detailedlearning}), explicitly indicating the training data used by each method. While several end-to-end approaches improve performance on natural-image compositional benchmarks such as \textsc{SugarCrepe} and \textsc{BiVLC}, gains on the \textsc{Swap} subsets—shown in brackets—remain limited, particularly on \textsc{BiVLC}. Their improvements also do not transfer to the controlled OOD benchmark \textsc{BiSCoR-Ctrl}. Despite being trained on additional compositional objectives or extended datasets (e.g., COCO variants with hard negatives or CC3M/12M), their performance on \textsc{BiSCoR-Ctrl} remains close to that of the original CLIP. 

In contrast, CLIP$\mathbf{+}$TF\textsubscript{Local} yields a substantial improvement on \textsc{BiSCoR-Ctrl} (from 1.4 to 15.1) without modifying the pretrained encoders. It also achieves stronger performance on the \textsc{Swap} subsets, particularly on \textsc{BiVLC}, reflecting improved binding under challenging perturbations. These gains stem from learning to align patch- and token-level embeddings over frozen representations.

Taken together with Section~\ref{subsec:main-results}, these results suggest that improved compositional generalization does not necessarily require altering large-scale representation learning. Rather, the way multimodal similarity is computed plays a critical role. Even models trained with compositional objectives show limited gains if inference relies solely on global embedding similarity, whereas learning localized alignment mechanisms leads to markedly stronger improvements under distribution shift. Notably, the magnitude of these gains varies across backbones: alignment learning over Perception Encoder yields substantially higher compositional performance than over CLIP, indicating that representational quality and training scale still influence the effectiveness of structured inference. While our findings highlight the central role of alignment mechanisms, they do not preclude that improved representation learning objectives may further enhance compositional robustness when combined with structured inference.

\section{Conclusion}
\label{sec:Conclusion}
In this work, we revisited the compositional limitations of dual-encoder VLMs. Rather than attributing these failures solely to deficient representations, we provided evidence that they are strongly influenced by the global cosine similarity typically used at inference time.

Through controlled diagnostic experiments, we demonstrated that enforcing structured region–segment alignment—while keeping pretrained encoders frozen—substantially improves compositional performance. Moreover, learning fine-grained alignment from frozen patch and token embeddings outperforms both the pretrained backbone and its fully fine-tuned counterpart, which rely on global cosine similarity at inference, particularly under distribution shift. In contrast, increasing model capacity alone by applying a transformer to global embeddings yields limited gains.  

These findings suggest that dual-encoder VLMs encode richer compositional information than commonly assumed, but standard global matching underutilizes it. We hope this work encourages renewed attention to inference protocols and evaluation practices when assessing compositional reasoning in VLMs.

\section*{Acknowledgements}

This work is partially supported by Ministry of Science, Innovation, and Universities of the Spanish Government MCIN/AEI/10.13039/501100011033 by means of the projects: (i) MOLVI (PID2024-157855OB-C32) and by FEDER, EU; (ii) HumanAIze (AIA2025-163322-C61). The CHIST-ERA grant (Project Geo-R2LLM, CHIST-ERA-23-MultiGIS-04) funded by the Ministry of Science, Innovation, and Universities of the Spanish Government (PCI2025-163286), the Basque Government (IXA excellence research group IT1570-22 and IKER-GAITU project), and the European Union under Horizon Europe (Project LUMINOUS, grant number 101135724).

\bibliographystyle{unsrtnat}
\bibliography{bibliography}

\appendix

\clearpage
\setcounter{page}{1}
\appendix
\onecolumn

\begin{center}

{\bf {\Large Appendix of Revisiting Compositionality in Dual-Encoder Vision-Language Models: The Role of Inference}} 
\end{center}

\section{\textsc{BiSCoR-Ctrl} dataset information}\label{appendix:dataset_info}

We host \textsc{BiSCoR} at HuggingFace\footnote{https://huggingface.co/datasets/imirandam/BiSCoR}. We provide a summary below.

\paragraph{Dataset documentation} \textsc{BiSCoR-Ctrl} is a benchmark of Bidirectional \textsc{Swaps} for Compositional Reasoning development.
Each instance consists of two images and two captions. Using each of the images and captions as a base, a model is asked to select the pair that correctly represents the base versus the hard negative distractor with minor compositional changes. Thus, we can measure image-to-text and text-to-image retrieval with hard negative pairs. To obtain good results on the dataset, it is necessary that the model performs well in both directions for the same instance.

\textsc{BiSCoR-Ctrl} is designed to have full control of the scenes and their compositions, allowing for a more detailed assessment of compositionality. It is based on CLEVR \citep{johnson2017clevr}, where we build different variants of \textsc{Swap} instances: i) \textsc{Color}, ii) \textsc{Size}, iii) \textsc{Material} and iv) \textsc{Quantity}. Each variant consists of two splits, Development and Test, with 1,000 instances each.

\begin{itemize}
    \item image: New positive image rendered by us.
    \item caption: Caption obtained from the scene used to render the positive image.
    \item negative\_image: New negative image rendered by us.
    \item negative\_caption: Caption obtained from the scene used to render the negative images.
\end{itemize}

\begin{figure}[h]
  \centering \includegraphics[scale=0.465]{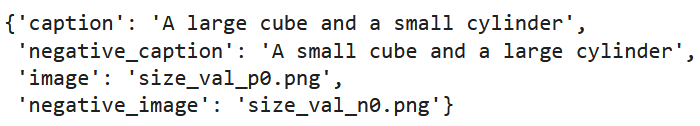}
  \caption{Example of a \textsc{BiSCoR} instance after loading the dataset.}
  \label{fig:instance_example}
\end{figure}

An example of an instance can be seen in Figure~\ref{fig:instance_example}. 

\paragraph{Maintenance plan} We are committed to maintaining the dataset to resolve any technical issues. We actively track issues in the HuggingFace or GitHub repositories. %
\paragraph{Licensing} Our work is licensed under the MIT License\footnote{https://github.com/IMirandaM/revisiting-vl-compositional-inference/blob/main/LICENSE} for the code and a Creative Commons Attribution 4.0 International License (CC BY 4.0) for the data\footnote{https://huggingface.co/datasets/choosealicense/licenses/blob/main/markdown/cc-by-4.0.md}. 
\paragraph{Author statement} We, the authors, assume full responsibility in case of violation of rights.

\section{Structure-Guided Inference development Results}
\label{appendix:dev_results}

In this section, we present the development experiments and results of Structure-Guided Inference (SGI) for CLIP, Perception Encoder and SigLIP 2 models.

\subsection{SGI development experiments}

We evaluate six different configurations of Structure-Guided Inference. For image crops, we always use crops of sizes (32, 32), (56, 56), (112, 112), (224, 224), (56, 112) and (112, 56), combining different scales and aspect ratios. We resize all the crops to the input size of the model and we deploy those crops in two different ways: i) \textit{grid}, avoiding any overlap of crops of the same size, and ii) \textit{overlap}, using a stride of $crop\_size / 2$. This means that we process 86 crops per image with the \textit{grid} configuration, and 270 crops with \textit{overlap}. 

\begin{figure*}[ht]
    \centering 
         \includegraphics[width=0.6\textwidth]{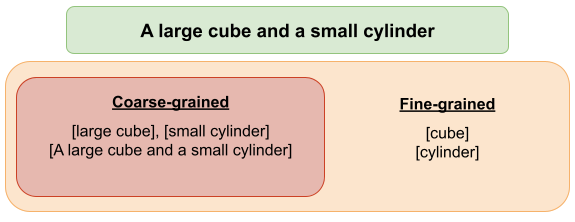}   
    \caption{An example for our two text segmenting strategies. As can be seen, \textit{fine-grained} adds two more segments to \textit{coarse-grained} text segments.}
    \label{fig:text-segments}   
\end{figure*}

Regarding text segmentation, we consider two different strategies: i) \textit{fine-grained}, where text segments are of the form [object], [attribute $\mathbf{+}$ object] and [attribute $\mathbf{+}$ object, relation, attribute $\mathbf{+}$ object] and ii) \textit{coarse-grained} text segments,  where text segments are of the form [attribute $\mathbf{+}$ object] and [attribute $\mathbf{+}$ object, relation, attribute $\mathbf{+}$ object] (see Figure~\ref{fig:text-segments}). To obtain those text segments in \textsc{BiSCoR-Ctrl} we directly use the ground-truth scene graphs, so we know that the created text segments are perfect. We have also used SpaCy to obtain the segments, see Appendix~\ref{appendix:sgi_spacy} for more details.

\subsection{CLIP development results}

Table~\ref{tab:dev-clip} shows the results of the development dataset for global embedding similarity and for different variants of Structure-Guided Inference for CLIP.

\begin{table*}[ht]
  \caption{Development results for CLIP base model and different SGI configurations. FG for \textit{fine-grained} and CG for \textit{coarse-grained} text segments. Bold for best results.}
  \label{tab:dev-clip}
  \centering
  \resizebox{1\width}{!}{%
 \begin{tabular}{p{2.5cm}cccccccccccccccc}
    \toprule
     \multirow{2}{*}{\small\textbf{Model}} &
      \multicolumn{12}{c}{\textbf{\textsc{BiSCoR-Ctrl}}} \\
      & \multicolumn{3}{c}{\textbf{\textsc{Color}}} & \multicolumn{3}{c}{\textbf{\textsc{Size}}} & \multicolumn{3}{c}{\textbf{\textsc{Material}}}& \multicolumn{3}{c}{\textbf{\textsc{Quantity}}} \\
\cmidrule(lr){2-4} \cmidrule(lr){5-7}  \cmidrule(lr){8-10}  \cmidrule(lr){11-13}  \cmidrule(lr){14-16} 
     &  \textbf{I2T} & \textbf{T2I} & \textbf{Gr.}&  \textbf{I2T} & \textbf{T2I} & \textbf{Gr.} &  \textbf{I2T} & \textbf{T2I} & \textbf{Gr.}&  \textbf{I2T} & \textbf{T2I} & \textbf{Gr.} \\
    \midrule    
    Random &  25.0 & 25.0 & 16.7 & 25.0 & 25.0 & 16.7 & 25.0 & 25.0 & 16.7 & 25.0 & 25.0 & 16.7\\
    \midrule
    CLIP  &  12.8                  & 5.3                   & 2.0                   & 2.8                   & 8.1                   & 0.4                   & 8.8                   & 2.7                   & 0.5                   & \textbf{11.9}         & 7.8                   & 3.5                   \\
    \midrule
    \quad Grid $\mathbf{+}$ FG &  70.1                  & 61.6                  & 53.8                  & 18.0                  & 9.0                   & 3.7                   & 23.6                  & 13.8                  & 6.9                   & 8.5                   & 11.5                  & 2.5                   \\
    \quad Grid $\mathbf{+}$ CG &   70.1                  & 67.4                  & 58.3                  & 18.0                  & 15.8                  & 6.6                   & 23.6                  & 16.6                  & 7.7                   & 6.9                   & 10.7                  & 2.1                   \\
    \midrule
    \quad Over $\mathbf{+}$ FG & \textbf{84.9}         & 75.0                  & 70.4                  & \textbf{20.0}         & 10.2                  & 4.5                   & \textbf{32.9}         & 19.4                  & 10.3                  & 10.2                  & \textbf{16.4}         & \textbf{4.7}          \\
    \quad Over $\mathbf{+}$ CG &   \textbf{84.9}         & \textbf{81.5}         & \textbf{75.6}         & \textbf{20.0}         & \textbf{16.2}         & \textbf{7.2}          & \textbf{32.9}         & \textbf{22.7}         & \textbf{12.9}         & 8.6                   & 14.5                  & 3.6                   \\   
 
    \bottomrule
  \end{tabular}}
\end{table*}

\subsection{Perception Encoder development results}
Table~\ref{tab:dev-pe} shows the results of the development dataset for global embedding similarity and for different variants of Structure-Guided Inference for Perception Encoder model.

\begin{table*}[ht]
  \caption{Development results for Perception Encoder base model and different SGI configurations. FG for \textit{fine-grained} and CG for \textit{coarse-grained} text segments. Bold for best results.}
  \label{tab:dev-pe}
  \centering
  \resizebox{1\width}{!}{%
 \begin{tabular}{p{2.5cm}cccccccccccccccc}
    \toprule
     \multirow{2}{*}{\small\textbf{Model}} &
      \multicolumn{12}{c}{\textbf{\textsc{BiSCoR-Ctrl}}} \\
      & \multicolumn{3}{c}{\textbf{\textsc{Color}}} & \multicolumn{3}{c}{\textbf{\textsc{Size}}} & \multicolumn{3}{c}{\textbf{\textsc{Material}}}& \multicolumn{3}{c}{\textbf{\textsc{Quantity}}} \\
\cmidrule(lr){2-4} \cmidrule(lr){5-7}  \cmidrule(lr){8-10}  \cmidrule(lr){11-13}  \cmidrule(lr){14-16} 
     &   \textbf{I2T} & \textbf{T2I} & \textbf{Gr.}&  \textbf{I2T} & \textbf{T2I} & \textbf{Gr.} &  \textbf{I2T} & \textbf{T2I} & \textbf{Gr.}&  \textbf{I2T} & \textbf{T2I} & \textbf{Gr.} \\
    \midrule    
    Random &  25.0 & 25.0 & 16.7 & 25.0 & 25.0 & 16.7 & 25.0 & 25.0 & 16.7 & 25.0 & 25.0 & 16.7\\
    \midrule
    PE  &  23.8                  & 9.5                   & 4.5                   & 36.0                  & 25.7                  & 11.4                  & 23.3                  & 12.7                  & 5.9                   & 24.8                  & 21.0                  & 11.8                  \\
    \midrule
    \quad Grid $\mathbf{+}$ FG &   91.2                  & 80.9                  & 78.2                  & 41.7                  & 15.1                  & 7.6                   & 46.2                  & 66.8                  & 35.8                  & 29.0                  & 26.6                  & 16.4                  \\
    \quad Grid $\mathbf{+}$ CG &    91.3                  & 85.2                  & 82.5                  & 43.1                  & 22.2                  & 11.8                  & 46.3                  & 72.3                  & 38.8                  & 23.9                  & 28.6                  & 14.0                  \\
    \midrule
    \quad Over $\mathbf{+}$ FG &  \textbf{96.8}         & 91.5                  & 90.3                  & 46.7                  & 20.6                  & 12.5                  & \textbf{51.5}         & 82.0                  & 47.0                  & \textbf{33.7}         & 30.9                  & \textbf{21.2}         \\
    \quad Over $\mathbf{+}$ CG &   \textbf{96.8}         & \textbf{94.2}         & \textbf{92.8}         & \textbf{47.5}         & \textbf{34.6}         & \textbf{21.3}         & 51.4                  & \textbf{85.9}         & \textbf{47.7}         & 26.2                  & \textbf{33.2}         & 16.9                  \\   
 
    \bottomrule
  \end{tabular}}
\end{table*}

\subsection{SigLIP 2 development results}

Table~\ref{tab:dev-siglip} shows the results of the development dataset for global embedding similarity and for different variants of Structure-Guided Inference for SigLIP 2 model.

\begin{table*}[ht]
  \caption{Development results for SigLIP 2 base model and different SGI configurations. FG for \textit{fine-grained} and CG for \textit{coarse-grained} text segments. Bold for best results.}
  \label{tab:dev-siglip}
  \centering
  \resizebox{1\width}{!}{%
 \begin{tabular}{p{2.5cm}cccccccccccccccc}
    \toprule
     \multirow{2}{*}{\small\textbf{Model}} &
      \multicolumn{12}{c}{\textbf{\textsc{BiSCoR-Ctrl}}} \\
      & \multicolumn{3}{c}{\textbf{\textsc{Color}}} & \multicolumn{3}{c}{\textbf{\textsc{Size}}} & \multicolumn{3}{c}{\textbf{\textsc{Material}}}& \multicolumn{3}{c}{\textbf{\textsc{Quantity}}} \\
\cmidrule(lr){2-4} \cmidrule(lr){5-7}  \cmidrule(lr){8-10}  \cmidrule(lr){11-13}  \cmidrule(lr){14-16} 
     &   \textbf{I2T} & \textbf{T2I} & \textbf{Gr.}&  \textbf{I2T} & \textbf{T2I} & \textbf{Gr.} &  \textbf{I2T} & \textbf{T2I} & \textbf{Gr.}&  \textbf{I2T} & \textbf{T2I} & \textbf{Gr.} \\
    \midrule      
    Random &  25.0 & 25.0 & 16.7 & 25.0 & 25.0 & 16.7 & 25.0 & 25.0 & 16.7 & 25.0 & 25.0 & 16.7\\
    \midrule
    SigLIP 2  &  17.1                  & 5.6                   & 2.4                   & 39.0                  & 16.0                  & 8.3                   & 6.7                   & 6.9                   & 1.0                   & 21.4                  & 10.4                  & 5.2                   \\
    \midrule
    \quad Grid $\mathbf{+}$ FG &  97.1                  & 89.6                  & 88.6                  & 17.7                  & 12.4                  & 4.1                   & 68.9                  & 63.4                  & 49.7                  & 35.0                  & 28.1                  & 17.7                  \\
    \quad Grid $\mathbf{+}$ CG &   97.1                  & 93.0                  & 92.0                  & 17.7                  & 14.7                  & 5.0                   & 68.9                  & 71.5                  & 55.5                  & 30.7                  & 28.0                  & 17.1                  \\
    \midrule
    \quad Over $\mathbf{+}$ FG & \textbf{82.0} &  95.3                  & 95.1                  & \textbf{19.4}         & 18.6                  & 5.8                   & \textbf{74.1}         & 78.4                  & 62.3                  & \textbf{37.3}         & \textbf{31.9}         & \textbf{21.3}         \\
    \quad Over $\mathbf{+}$ CG &    \textbf{98.6}         & \textbf{96.1}         & \textbf{95.9}         & \textbf{19.4}         & \textbf{21.4}         & \textbf{5.9}          & \textbf{74.1}         & \textbf{82.7}         & \textbf{64.9}         & 32.6                  & 30.6                  & 18.6                  \\

    \bottomrule
  \end{tabular}}
\end{table*}

\subsection{SGI SpaCy results}\label{appendix:sgi_spacy}

To evaluate the robustness of SGI, we transitioned from manual ground-truth labels to a simplified, automated extraction pipeline using the SpaCy~\citep{spacy} noun chunks utility. This shift was specifically designed to determine whether the performance gains were a result of a better exploitation of pretrained representations or merely a reflection of high-fidelity text segments. By employing a standard, non-LLM parser (i.e. "en\_core\_web\_sm") to decompose captions into entity-attribute pairs (e.g., a red cube) and global captions (e.g., a red cube and a blue sphere), allows us to verify if SGI can maintain its efficacy when decoupled from gold annotations. Table~\ref{tab:sgi_spacy} shows the results for CLIP, PE, and SigLIP2 with SpaCy text segments, which are similar to those obtained with the gold annotations.

\begin{table}[t]
  \caption{Results for SGI with text segments obtained with SpaCy on the \textsc{BiSCoR-Ctrl} dataset. We provide the average group score obtained by each model using global embedding similarity, SGI with gold annotations and SGI with SpaCy text segments. We also provide the group score for all the categories of the dataset.}  
  \label{tab:sgi_spacy}
  \centering  
  \begin{tabular}{p{2cm}p{1.5cm}cccc}
    \toprule
    \multirow{2}{*}{\textbf{Model}} & \multirow{2}{1.5cm}{\centering \textbf{Average}} & \multicolumn{4}{c}{\textbf{Categorie}s} \\
     &  & \textsc{\textbf{Color}} & \textsc{\textbf{Size}} & \textsc{\textbf{Material}} & \textsc{\textbf{Quantity}} \\
    \midrule
    CLIP & \centering 1.4 & 1.5 & 0.2 & 0.7 & 3.3 \\
    \quad $\mathbf{+}$SGI & \centering24.9 & \textbf{76.5} & 4.7 & \centering15.7 & 2.5 \\
    \quad $\mathbf{+}$SGI\textsubscript{SpaCy} & \centering\textbf{25.5}& 74.9 & \textbf{5.1} & \textbf{18.0} & \textbf{3.9} \\
    \midrule
    PE & \centering8.5 & 5.9 & 10.9 & 7.7 & 9.4 \\
    \quad $\mathbf{+}$SGI & \centering\textbf{44.7} & 91.3 & \textbf{16.9} & 55.5 & \textbf{15.3} \\
    \quad $\mathbf{+}$SGI\textsubscript{SpaCy} & \centering\textbf{44.7} & \textbf{91.5} & 12.3 & \textbf{60.8} & 14.1 \\
    \midrule
    SigLIP 2 & \centering4.5 & 3.7 & \textbf{9.6} & 1.5 & 3.1 \\
    \quad $\mathbf{+}$SGI & \centering\textbf{47.6} & \textbf{94.9} & 7.1 & \textbf{68.3} & \textbf{20.2} \\
    \quad $\mathbf{+}$SGI\textsubscript{SpaCy} &\centering45.7 &94.4&3.6 & \textbf{68.3}& 16.5\\
    \bottomrule
  \end{tabular}
\end{table}

\section{Detailed evaluation metrics}
\label{appendix:metrics}
The \textbf{I2T} score measures the performance for image-to-text retrieval. For each instance in development and test datasets, we actually have two image-to-text retrieval examples. To obtain a perfect I2T score, the correct captions for both images have to be selected. Thus, assuming $C_0, C_1$ refer to positive and negative caption respectively, $I_0, I_1$ to positive and negative image, and we use $s(C_i, I_i)$ as the similarity function for a caption and an image, I2T score $I2T(C_0, I_0, C_1, I_1)$ is defined in Equation \ref{eq:text-score}: 
     \begin{equation}
     \label{eq:text-score}
    I2T\left(C_{0}, I_{0}, C_{1}, I_{1}\right)=\left\{\begin{array}{cc}
     1 & \textnormal { if } s\left(C_{0}, I_{0}\right)>s\left(C_{1}, I_{0}\right)\\
    & \textnormal { and } s\left(C_{1}, I_{1}\right)>s\left(C_{0}, I_{1}\right) \\
     0 & \textnormal { otherwise }
    \end{array}\right.
    \end{equation}
    
The \textbf{T2I} score $T2I(C_{0}, I_{0}, C_{1}, I_{1})$ is similarly defined for text-to-image retrieval (Equation \ref{eq:img-score}):

    \begin{equation}
    \label{eq:img-score}
    T2I\left(C_{0}, I_{0}, C_{1}, I_{1}\right)=\left\{\begin{array}{cc}
     1 & \textnormal { if } s\left(C_{0}, I_{0}\right)>s\left(C_{0}, I_{1}\right)  \\
    & \textnormal { and } s\left(C_{1}, I_{1}\right)>s\left(C_{1}, I_{0}\right) \\
     0 & \textnormal { otherwise }
    \end{array}\right.
    \end{equation}

Finally, the \textbf{Group} score $G(C_{0}, I_{0}, C_{1}, I_{1})$ is the main metric, since it combines the performance for image-to-text and text-to-image retrieval. To obtain a perfect group score for a given instance, both images have to be matched with the suitable captions and both captions with the suitable images. The group score is defined in Equation \ref{eq:group-score}:

    \begin{equation}    
    \label{eq:group-score}
    G\left(C_{0}, I_{0}, C_{1}, I_{1}\right)=\left\{\begin{array}{cc}
    1 & \textnormal { if } I2T\left(C_{0}, I_{0}, C_{1}, I_{1}\right)\\
    & \textnormal { and } T2I\left(C_{0}, I_{0}, C_{1}, I_{1}\right) \\
    0 & \textnormal { otherwise }
    \end{array}\right.
    \end{equation}

\section{Learning Fine-Grained Alignment development}

In this section you can find the information about the training hyperparameters and development results of Section~\ref{sec:learning} of the main paper.

\subsection{Training hyperparameters}
\label{appendix:learning_setup}

Hyperparameters for TF\textsubscript{Global} and TF\textsubscript{Local} training:
\begin{itemize}
    \item Learning rate: 1e-4.
    \item Scheduler: Cosine scheduler with a warm-up of 10\% of the total number of training steps.
    \item Optimizer: AdamW optimizer with beta1 = 0.9, beta2 = 0.95, eps = 1e-08 and weight decay = 1e-7.
    \item Loss function: Contrastive Loss.
    \item Batch size: We define a batch size of 50 (50 images x 50 captions) for COCO dataset. For TROHN-Img, as we have hard negatives, we define a batch size of 25, and then we add negatives, obtaining 25 positive pairs and 25 negative pairs in the same batch.
    \item Epochs: We fine-tune all models over 5 epochs and we use validation accuracy as the model selection criterion, i.e. we selected the model with the highest accuracy on the corresponding validation set.
    \item Embedding layer: For TF\textsubscript{Global}, because it relies on the same embeddings used for global embedding similarity, we extract representations from the final layer. For CLIP TF\textsubscript{Local}, following the established practice in the literature, e.g. LLaVA \citep{liu2023visual} and VQAScore \citep{lin2024revisiting}, we use embeddings from the penultimate layer for both encoders, vision and text. For PE TF\textsubscript{Local} we use embeddings from the last layer.  
\end{itemize}

For all PE full fine-tunings we have used the same hyperparameters as in \citep{miranda2024BiVLC} for CLIP\textsubscript{COCO} and CLIP\textsubscript{TROHN-Img}. Detailed hyperparameters:

\begin{itemize}
    \item Learning rate: 1e-6.
    \item Scheduler: Cosine scheduler with 50 warmup steps.
    \item Optimizer: AdamW optimizer with beta1 = 0.9, beta2 = 0.98, eps = 1e-6 and weight decay = 0.1.
    \item Loss function: InfoNCE Loss.
    \item Batch size: We define a batch size of 400 (400 images x 400 captions) with COCO dataset and for TROHN-Img a batch size of 200, and then we add the 200 negatives, resulting in 400 images x 400 captions.
    \item Epochs: We fine-tune all models over 10 epochs and we used validation accuracy as the model selection criterion, i.e. we selected the model with the highest accuracy on the corresponding validation set.
\end{itemize}

More information about training data, models, software and hardware can be found in Appendix~\ref{appendix:implementation}

\subsection{Learning Fine-Grained Alignment development results}\label{appendix:layer_setup}

Table~\ref{tab:transformer-dev} shows the results of our four learnable variants with different number of layers, from 1 to 4, in the validation set of each training data.

\begin{table*}[t]
  \caption{Development results for each variant and dataset with different number of layers. For validation, we provide batch accuracy. In bold, the best for each variant. 
  }
  \label{tab:transformer-dev}
  \centering
  \resizebox{0.89\width}{!}{%
  \begin{tabular}{p{1.5cm}p{1.5cm}ccccc}
    \toprule
    \multirow{2}{*}{\small\textbf{Model}} & \multirow{2}{*}{\small\textbf{Dataset}} & \multirow{2}{*}{\small\textbf{Nº layers}} & \textbf{\textsc{Validation}} \\
     \cmidrule(lr){4-4} \cmidrule(lr){5-7}  
     &  & & \textbf{Accuracy} \\
    \midrule    
    \multirow{8}{*}{TF\textsubscript{Global}} &\multirow{4}{*}{COCO}  & 1 & 91.7            \\
    & & 2 & 93.0          \\
    & & 3 & \textbf{93.2}    \\
     & & 4 & \textbf{93.2}         \\
    \cmidrule(lr){2-7} 
    & \multirow{4}{*}{TROHN-Img} & 1 & 80.5                     \\
    & & 2 & 83.8                    \\
    & & 3 & 84.0           \\
     & & 4 & \textbf{84.3}                  \\
    \midrule
   \multirow{8}{*}{TF\textsubscript{Local}} &\multirow{4}{*}{COCO}  & 1 & 91.5           \\
    & & 2 & 94.2       \\
    & & 3 & 94.3            \\
     & & 4 & \textbf{94.4}         \\
    \cmidrule(lr){2-7} 
    & \multirow{4}{*}{TROHN-Img} &  1 & 81.8               \\
    & & 2 & 86.4                 \\
    & & 3 & 86.8                \\
     & & 4 & \textbf{87.2}           \\
    
    \bottomrule
  \end{tabular}}
\end{table*}

\section{Implementation details}\label{appendix:implementation}

This appendix contains all the information related to the implementation of the experiments. All the source can be found at \url{https://github.com/IMirandaM/revisiting-vl-compositional-inference}. %

\subsection{Source datasets}\label{appendix:source}
We obtain all source datasets directly from the original sources published by the authors. To the best of our knowledge, all data sources we use are open to non-commercial use, do not contain personally identifiable information and do not contain offensive content.

\begin{itemize}
    \item \textbf{CLEVR} \citep{lin2014microsoft}: We obtain CLEVR scenes from the official project website\footnote{\url{https://cs.stanford.edu/people/jcjohns/clevr/}} under a Creative Commons Attribution 4.0 License.
\end{itemize}

\subsection{Training datasets}\label{appendix:training_data}
To train TF\textsubscript{Global} and TF\textsubscript{Local}, we used the following two datasets: 

\begin{itemize}
    \item \textbf{COCO} \citep{lin2014microsoft}: We obtain COCO 2017 from the official project website\footnote{\url{https://cocodataset.org/\#download}} under a Creative Commons Attribution 4.0 License\footnote{\url{https://cocodataset.org/\#termsofuse}}. It contains 591,753 captions and 118,287 images, i.e., 591,753 instances formed by an image and a caption.
    \item \textbf{TROHN-Img} \citep{miranda2024BiVLC}: We obtain TROHN-Img from the official Hugging Face repository \footnote{\url{https://huggingface.co/datasets/imirandam/TROHN-Img}} under the MIT license. It contains 296,070 instances formed by two images and two captions, i.e. 592,140 pairs, an amount similar to that of the COCO 2017 train. 
\end{itemize}

\subsection{Evaluation datasets}\label{appendix:eval_data}

We obtain all evaluation datasets directly from the original sources published by the authors.

\begin{itemize}
    \item \textbf{\textsc{BiVLC}} \citep{miranda2024BiVLC}: We obtain \textsc{BiVLC} from the official Hugging Face repository\footnote{\url{https://huggingface.co/datasets/imirandam/BiVLC}}. 
    \item \textbf{\textsc{SugarCrepe}} \citep{hsieh2024sugarcrepe}: We obtain \textsc{SugarCrepe} from the official GitHub repository\footnote{\url{https://github.com/RAIVNLab/sugar-crepe}}.
\end{itemize}

\subsection{Software information}\label{appendix:software}
\paragraph{Models}
We detail the sources of models we used.

\begin{itemize}
    \item \textbf{CLIP:} We obtain the pretrained baseline VIT-B-32 OpenAI's CLIP model \citep{radford2021learning} from Hugging Face\footnote{\url{https://huggingface.co/openai/clip-vit-base-patch32}}.
   
    \item \textbf{SigLIP 2:} We obtain all SigLIP 2 \citep{tschannen2025siglip} models from Hugging Face collection\footnote{\url{https://huggingface.co/collections/google/siglip2-67b5dcef38c175486e240107}}.
    \begin{itemize}
        \item \textbf{SigLIP 2:} We obtain
siglip2-base-patch32-256 from the official Hugging Face repository\footnote{\url{https://huggingface.co/google/siglip2-base-patch32-256}}.
        \item \textbf{SigLIP 2-Giant:} We obtain
siglip2-giant-opt-patch16-256 from the official Hugging Face repository\footnote{\url{https://huggingface.co/google/siglip2-giant-opt-patch16-256}}.
    \end{itemize}
    \item \textbf{Perception Encoder:} We obtain all Perception Encoder \citep{bolya2025perception} models from Hugging Face collection\footnote{\url{https://huggingface.co/collections/facebook/perception-encoder}}.
    \begin{itemize}
        \item \textbf{Pe:} We obtain PE-Core-B16-224 from the official Hugging Face repository\footnote{\url{https://huggingface.co/facebook/PE-Core-B16-224}}.
        \item \textbf{PE-Giant:} We obtain PE-Core-G14-448 from the official Hugging Face repository\footnote{\url{https://huggingface.co/facebook/PE-Core-G14-448}}.
    \end{itemize}
    
    \item \textbf{NegCLIP:} We obtain the NegCLIP model \citep{yuksekgonul2022and} from the official GitHub repository.\footnote{\url{https://github.com/mertyg/vision-language-models-are-bows}}
    
    \item \textbf{TripletCLIP:} We obtain the TripletCLIP model \citep{patel2024TripletCLIP} from the official GitHub repository.\footnote{\url{https://github.com/tripletclip/TripletCLIP/}}
    \item \textbf{FSC-CLIP:} We obtain the FSC-CLIP model \citep{oh2024preserving} from the official GitHub repository.\footnote{\url{https://github.com/ytaek-oh/fsc-clip}}
    \item \textbf{FineCLIP:} We obtain the FineCLIP model \citep{jing2024fineclip} from the official GitHub repository.\footnote{\url{https://github.com/Timsty1/FineCLIP}}
    \item \textbf{X-VLM:} We obtain the X-VLM model fine-tuned for retrieval in COCO \citep{zeng2022multi} from the official GitHub repository.\footnote{\url{https://github.com/zengyan-97/X-VLM}}
    \item \textbf{CLIP\textsubscript{COCO}:} We obtain the CLIP\textsubscript{COCO} model \citep{miranda2024BiVLC} from the official Hugging Face repository.\footnote{\url{https://huggingface.co/imirandam/CLIP_COCO}}
    \item \textbf{CLIP\textsubscript{TROHN-Img}:} We obtain the CLIP\textsubscript{TROHN-Img} model \citep{miranda2024BiVLC} from the official Hugging Face repository.\footnote{\url{https://huggingface.co/imirandam/CLIP_TROHN-Img}}
\end{itemize}

\paragraph{Implementation decisions}

We have decided to keep the preprocessing of the images constant based on the model with the lowest resolution, i.e. CLIP, controlling that all models receive the same original image and same crops. For this, all images are preprocessed in the same way, resize to 224 and center crop. For SGI all hyperparameters described in appendix~\ref{appendix:dev_results} are identical across all models. The only difference is the batch size: 25 for all the models, except 10 for FineCLIP and X-VLM and and 5 for SigLIP 2-Giant. The exact values of all the hyperparameters for full fine-tunings, TF\textsubscript{Global} and TF\textsubscript{Local} can be found in Appendix~\ref{appendix:learning_setup}

\paragraph{Evaluation:} We base our evaluations on the Transformers library \citep{wolf-etal-2020-transformers}, except for X-VLM\footnote{\url{https://github.com/zengyan-97/X-VLM}}, FineCLIP\footnote{\url{https://github.com/Timsty1/FineCLIP}} and PE\footnote{\url{https://github.com/facebookresearch/perception_models}} where we use the code provided in the official repository of each project.

\paragraph{Rendering images:} For rendering the images from \textsc{BiSCoR-Ctrl} we have used Blender 4.4.3.

\subsection{Hardware information}\label{appendix:hardware}

\paragraph{Development experiments:} All development experiments have been performed on one NVIDIA A100-SXM4-80GB GPU and 64 GB of RAM.

\paragraph{Evaluation:} As in the development experiments, the evaluation was performed on one NVIDIA A100-SXM4-80GB GPU and 64 GB of RAM.

\paragraph{Rendering images:} For rendering the images we have used an NVIDIA RTX A1000 6GB Laptop GPU. Each rendering takes around 1.5 seconds. 

\section{Detailed results of Learning Fine-Grained Alignment}\label{appendix:detailedlearning}

In this appendix you can find the detailed results of all the variants for each model, CLIP and PE, and the results for end-to-end models in the three test datasets.

\subsection{\textsc{SugarCrepe} detailed results}\label{appendix:sugarcrepe_detailed}

Table~\ref{tab:sugarcrepe_detailed} shows the results  of all the variants for each model, CLIP and PE, and the results for end-to-end models in \textsc{SugarCrepe}.

\begin{table*}
  \caption{Average group score and per category for  \textsc{SugarCrepe} divided in CLIP and PE families. In bold, the best for each family and underlined the second best.}
 \label{tab:sugarcrepe_detailed}
  \centering  
  \resizebox{1\width}{!}{%
  \begin{tabular}{lcccc}%
    \toprule    
      \multirow{2}{*}{\small\textbf{Model}}  & \multicolumn{4}{c}{\textbf{\textsc{SugarCrepe}}}\\
      
    &\textbf{\textsc{Average}}&\multicolumn{1}{c}{\textbf{\textsc{Replace}}}&\multicolumn{1}{c}{\textbf{\textsc{Swap}}}&\multicolumn{1}{c}{\textbf{\textsc{Add}}}\\
    \midrule
    CLIP & 73.0  & 80.8 & 63.3 & 75.1 \\
     \quad FT\textsubscript{COCO} & 80.9  & 86.1	& 68.9	& 87.8  \\    
    \quad FT\textsubscript{TROHN-img} & \underline{85.5}  & \textbf{89.4}	& 72.0 &	\textbf{95.2}  \\  
    \quad TF\textsubscript{Global-COCO} & 73.2  &  80.7	& 63.5 &	75.5  \\ 
    \quad TF\textsubscript{Global-TROHN-img} & 82.2  &  85.3 &	68.8 &	92.5  \\  
    \quad TF\textsubscript{Local-COCO} & 80.9  & 83.2	& 76.3	& 83.2  \\    
    \quad TF\textsubscript{Local-TROHN-img} &  \textbf{86.3} &  86.9	& \underline{77.1} &	\underline{94.8}  \\ 
    NegCLIP &  83.6 & 85.5	& 76.6 & 88.8 \\
    TripletCLIP & 82.6  & \underline{88.8}	& 71.9	& 87.2    \\ 
    FSC-CLIP & 85.1 & 88.1	& \textbf{77.6}	& 89.5   \\ 
    X-VLM & 81.9 & 88.2	& 67.4	& 90.1   \\ 
    FineCLIP & 80.6  &  86.0	&  68.2	&  87.7 \\ 
    \midrule
    PE & 84.4 & 87.9 &	76.8	& 88.4 \\
    \quad FT\textsubscript{COCO} & 85.7  & 88.2	& 76.1 &	92.9   \\    
    \quad FT\textsubscript{TROHN-img} & \textbf{90.1}  & \textbf{92.4}	& 80.3 &	\textbf{97.5}  \\  
    \quad TF\textsubscript{Global-COCO} & 78.0  &  84.5	& 68.0 &	81.4  \\ 
    \quad TF\textsubscript{Global-TROHN-img} & 86.9  & 89.9	& 74.9	& \underline{95.9}  \\  
    \quad TF\textsubscript{Local-COCO} & 84.4  & 85.2	& \underline{80.6}	& 87.4  \\    
    \quad TF\textsubscript{Local-TROHN-img} & \underline{89.2}  & \underline{91.3}	& \textbf{80.8} &	95.6 \\

    \bottomrule
  \end{tabular}}
\end{table*}

\subsection{\textsc{BiVLC} detailed results}\label{appendix:bivlc_detailed}

Table~\ref{tab:bivlc_detailed} shows the results  of all the variants for each model, CLIP and PE, and the results for end-to-end models in \textsc{BiVLC}.

\begin{table*}
  \caption{Average group score and per category for  \textsc{BiVLC} divided in CLIP and PE families. In bold, the best for each family and underlined the second best.}
 \label{tab:bivlc_detailed}
  \centering  
  \resizebox{1\width}{!}{%
  \begin{tabular}{lcccc}%
    \toprule    
      \multirow{2}{*}{\small\textbf{Model}}  & \multicolumn{4}{c}{\textbf{\textsc{BiVLC}}}\\
      
    &\textbf{\textsc{Average}}&\multicolumn{1}{c}{\textbf{\textsc{Replace}}}&\multicolumn{1}{c}{\textbf{\textsc{Swap}}}&\multicolumn{1}{c}{\textbf{\textsc{Add}}}\\
    \midrule
    CLIP & 36.8  & 57.3	& 13.7 &	39.6 \\
     \quad FT\textsubscript{COCO} & 47.5  & 69.9	&20.9&	51.6  \\    
    \quad FT\textsubscript{TROHN-img} & \underline{57.5}  & \textbf{76.6}	&\underline{27.9}&	\underline{68.0}  \\  
    \quad TF\textsubscript{Global-COCO} & 38.4  &  59.9	&11.4&	44.0  \\ 
    \quad TF\textsubscript{Global-TROHN-img} & 49.0  &  69.1	& 15.9 &	61.9  \\  
    \quad TF\textsubscript{Local-COCO} & 45.7  & 62.8	&24.0 &	50.3  \\    
    \quad TF\textsubscript{Local-TROHN-img} &  \textbf{61.3} &  \underline{74.1}	& \textbf{39.0} &	\textbf{70.7}  \\ 
    NegCLIP &  44.9 & 68.0	& 18.7 &	48.0 \\
    TripletCLIP & 35.2  & 54.9	& 9.8 &	40.8    \\ 
    FSC-CLIP & 46.5 & 68.8	& 19.2 &	51.6   \\ 
    X-VLM & 40.9 & 64.1	&12.3&	46.3   \\ 
    FineCLIP & 39.4  &  63.5	& 8.6	& 46.1 \\ 
    \midrule
    PE & 41.5 & 64.8	& 13.4 & 46.3 \\
    \quad FT\textsubscript{COCO} & 56.3  & 75.8	&31.8&	61.3  \\    
    \quad FT\textsubscript{TROHN-img} & \textbf{68.2}  & \textbf{82.2}	& \textbf{43.7} &	\textbf{78.7}  \\  
    \quad TF\textsubscript{Global-COCO} & 44.6  &  66.8	& 15.6	& 51.4  \\ 
    \quad TF\textsubscript{Global-TROHN-img} & 58.3  & 77.5	& 24.5 &	72.8  \\  
    \quad TF\textsubscript{Local-COCO} & 53.3  & 69.7	& 33.7 &	56.4  \\    
    \quad TF\textsubscript{Local-TROHN-img} & \underline{67.1}  & \underline{80.9}	& \underline{42.9} &	\underline{77.7} \\

    \bottomrule
  \end{tabular}}
\end{table*}

\subsection{\textsc{BiSCoR-Ctrl} Detailed results}\label{appendix:BISCOR_detailed}

Table~\ref{tab:BISCOR_detailed} shows the results  of all the variants for each model, CLIP and PE, and the results for end-to-end models in \textsc{BiSCoR-Ctrl}.

\begin{table*}
  \caption{Average group score and per category for  \textsc{BiSCoR-Ctrl} divided in CLIP and PE families. In bold, the best for each family and underlined the second best.}
 \label{tab:BISCOR_detailed}
  \centering  
  \resizebox{1\width}{!}{%
  \begin{tabular}{lccccc}%
    \toprule    
      \multirow{2}{*}{\small\textbf{Model}}  & \multicolumn{5}{c}{\textbf{\textsc{BiSCoR-Ctrl}}}\\
      
    &\textbf{\textsc{Average}}&\multicolumn{1}{c}{\textbf{\textsc{Color}}}&\multicolumn{1}{c}{\textbf{\textsc{Size}}}&\multicolumn{1}{c}{\textbf{\textsc{Material}}}&\multicolumn{1}{c}{\textbf{\textsc{Quantity}}}\\
    \midrule
    CLIP & 1.4	& 1.5	& 0.2	& \underline{0.7}	& 3.3 \\
     \quad FT\textsubscript{COCO} & 1.4	& 1.9	& 1.3 &0.4	&1.9  \\    
    \quad FT\textsubscript{TROHN-img} & 1.9	& 5.1	&0.0 &	0.0&	2.3  \\  
    \quad TF\textsubscript{Global-COCO} & 1.2 &	1.8 &	1.4 &	0.2 &	1.2  \\ 
    \quad TF\textsubscript{Global-TROHN-img} & 1.2	&1.8&	0.0&	\textbf{1.6}&	1.4  \\  
    \quad TF\textsubscript{Local-COCO} & \textbf{15.1}	&\textbf{49.5}&	\textbf{1.6}&	0.1&	\textbf{9.2}  \\    
    \quad TF\textsubscript{Local-TROHN-img} &  \underline{13.2}	&\underline{46.4}	&0.1&	0.4&	\underline{5.8}  \\ 
    NegCLIP &  1.8	&2.4&	0.1&	0.5&	4.2 \\
    TripletCLIP & 1.2	&1.4&	0.7&	\underline{0.7}&	2.0    \\ 
    FSC-CLIP & 1.2	&1.6&	0.8&	0.0&	2.5   \\ 
    X-VLM & 1.7	&3.7&	0.0&	0.1&	3.1   \\ 
    FineCLIP & 1.4	&2.2&	\underline{1.5}&	0.2&	1.8 \\ 
    \midrule
    PE & 8.5	&5.9	&\underline{10.9}&	7.7&	9.4 \\
    \quad FT\textsubscript{COCO} & 8.3 &6.8 & 9.9 & 8.9 & 7.7 \\    
    \quad FT\textsubscript{TROHN-img} & 9.8 &9.7 & 7.9 & 10.0 & \underline{11.5}\\  
    \quad TF\textsubscript{Global-COCO} & 1.9	& 1.5 &	1.6 & 2.1 &	2.4  \\ 
    \quad TF\textsubscript{Global-TROHN-img} & 1.2	& 1.2 &	1.4 &	0.9 &	1.3 \\  
    \quad TF\textsubscript{Local-COCO} & \textbf{30.0}	& \textbf{72.4} &	\textbf{19.8} &	\underline{15.8} &	\textbf{12.1}  \\    
    \quad TF\textsubscript{Local-TROHN-img} & \underline{24.0}	&\underline{60.4} &	8.6 &	\textbf{20.5} &	6.5 \\

    \bottomrule
  \end{tabular}}
\end{table*}

\end{document}